\documentclass[10pt,twocolumn,letterpaper]{article}
\usepackage{booktabs}
\usepackage{multirow}
\usepackage{cvpr}
\usepackage{times}
\usepackage{epsfig}
\usepackage{graphicx}
\usepackage{amsmath}
\usepackage{amssymb}
\usepackage[utf8]{inputenc}

\usepackage[pagebackref=true,breaklinks=true,letterpaper=true,colorlinks,bookmarks=false]{hyperref}

\cvprfinalcopy 

\def\cvprPaperID{****} 

\ifcvprfinal\fi
\begin{document}

\title{Faster R-CNN Features for Instance Search}

\author{Amaia Salvador, Xavier Gir\'o-i-Nieto, Ferran Marqu\'es \\
Universitat Politècnica de Catalunya (UPC) \\
Barcelona, Spain \\
{\tt\small \{amaia.salvador,xavier.giro\}@upc.edu}
\and
Shin'ichi Satoh\\
National Institute of Informatics\\
Tokyo, Japan\\
{\tt\small satoh@nii.ac.jp}
}

\maketitle

\begin{abstract}

Image representations derived from pre-trained Convolutional Neural Networks (CNNs) have become the new state of the art in computer vision tasks such as instance retrieval. This work explores the suitability for instance retrieval of image- and region-wise representations pooled from an object detection CNN such as Faster R-CNN. We take advantage of the object proposals learned by a Region Proposal Network (RPN) and their associated CNN features to build an instance search pipeline composed of a first filtering stage followed by a spatial reranking. We further investigate the suitability of Faster R-CNN features when the network is fine-tuned for the same objects one wants to retrieve. We assess the performance of our proposed system with the Oxford Buildings 5k, Paris Buildings 6k and a subset of TRECVid Instance Search 2013, achieving competitive results.
\end{abstract}

\section{Introduction}
\label{introduction}

Visual media is nowadays the most common type of content in social media channels, thanks to the proliferation of ubiquitous cameras. This explosion of online visual content has motivated researchers to come up with effective yet efficient automatic content based image retrieval systems. This work addresses the problem of instance search, understood as the task of retrieving those images from a database that contain an instance of a query.

Recently, Convolutional Neural Networks (CNNs) have been proven to achieve state of the art performance in many computer vision tasks such as image classification \cite{alexnet,vgg}, object detection \cite{fasterrcnn} or semantic segmentation \cite{long2015fully}. CNNs trained with large amounts of data have been shown to learn feature representations that can be generic enough to be used even to solve tasks for which they had not been trained \cite{cnnofftheshelf}. Particularly for image retrieval, many works in the literature \cite{babenko2015,tolias2015,kalantidis2015} have adopted solutions based on off-the-shelf features extracted from a CNN pretrained for the task of image classification \cite{alexnet,vgg,googlenet}, achieving state of the art performance in popular retrieval benchmarks.

\begin{figure}
  \includegraphics[width=\columnwidth]{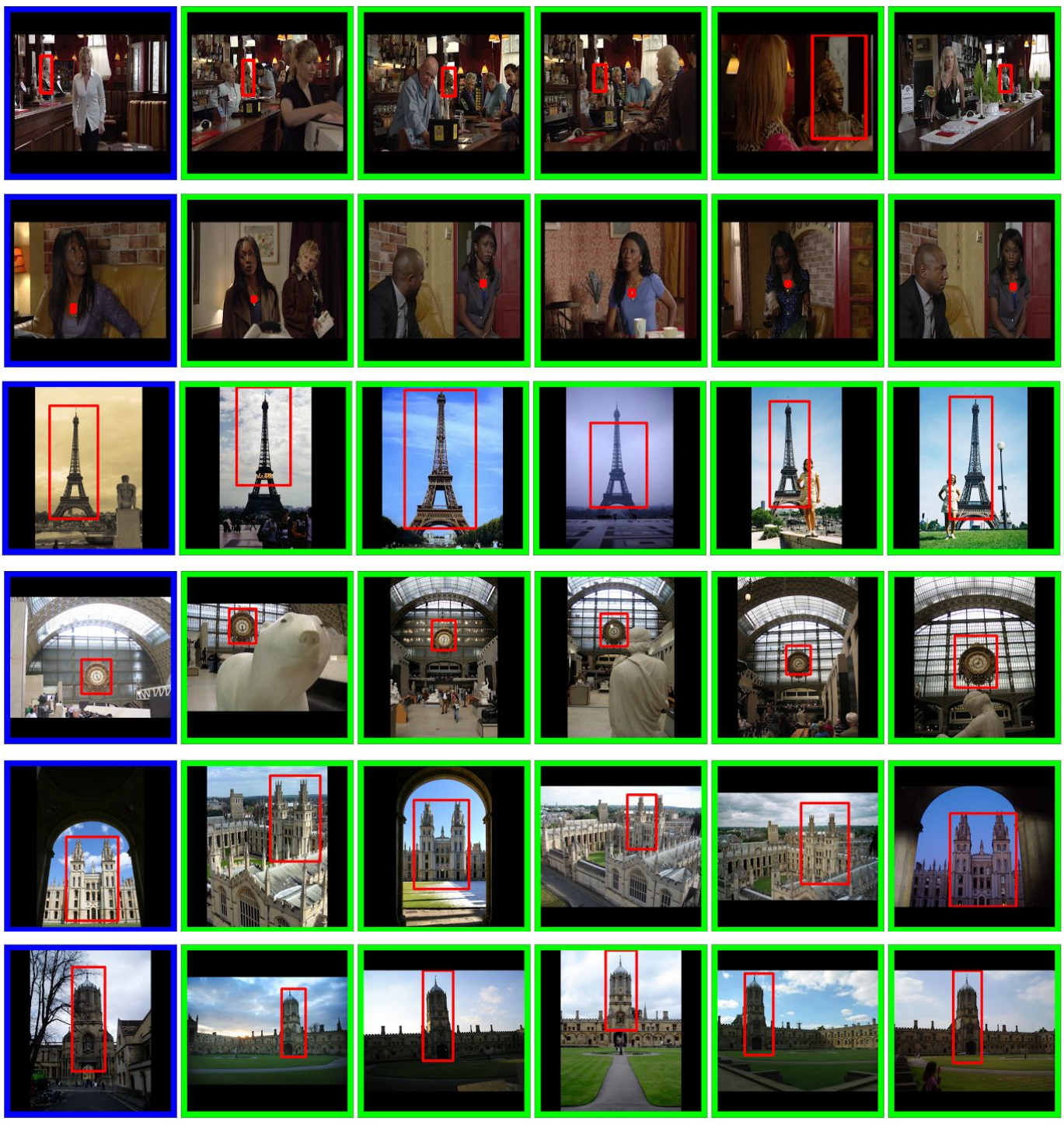}
  \caption{Examples of the rankings and object locations obtained by our proposed retrieval system for query objects (left, depicted with a blue contour) of three different datasets: TRECVid INS 2013, Paris Buildings and Oxford Buildings.}
  \label{res_examples}
\end{figure}

Instance search systems often combine fast first filtering stages, in which all images in a database are ranked according to their similarity to the query, with more computationally expensive mechanisms that are only applied to the top retrieved items. Geometric verification and spatial analysis ~\cite{jegou2010improving, zhang2011image, mei2014multimedia,zhang2015topological} are common reranking strategies, which are often followed with query expansion (pseudo-relevance feedback) ~\cite{arandjelovic2012three,chum2007total}.

Spatial reranking usually involves the usage of sliding windows at different scales and aspect ratios over an image. Each window is then compared to the query instance in order to find the optimal location that contains the query, which requires the computation of a visual descriptor on each of the considered windows. Such strategy resembles that of an object detection algorithm, which usually evaluates many image locations and determines whether they contain the object or not. Object Detection CNNs \cite{rcnn,sppnet,fastrcnn,fasterrcnn} have rapidly evolved to a point where the usage of exhaustive search with sliding windows or the computation of object proposals \cite{selectivesearch,mcg} is no longer required. Instead, state of the art detection CNNs \cite{fasterrcnn} are trained in an end-to-end manner to simultaneously learn object locations and labels. 

This work explores the suitability of both off-the-shelf and fine-tuned features from an object detection CNN for the task of instance retrieval. We make the following three contributions:

\begin{itemize}

\item We propose to use a CNN pre-trained for object detection to extract convolutional features both at global and local scale in a single forward pass of the image through the network.

\item We explore simple spatial reranking strategies, which take advantage of the locations learned by a Region Proposal Network (RPN) to provide a rough object localization for the top retrieved images of the ranking.

\item We analyze the impact of fine-tuning an object detection CNN for the same instances one wants to query in the future. We find such a strategy to be suitable for learning better image representations.

\end{itemize}

\begin{figure*}
  \includegraphics[width=\textwidth]{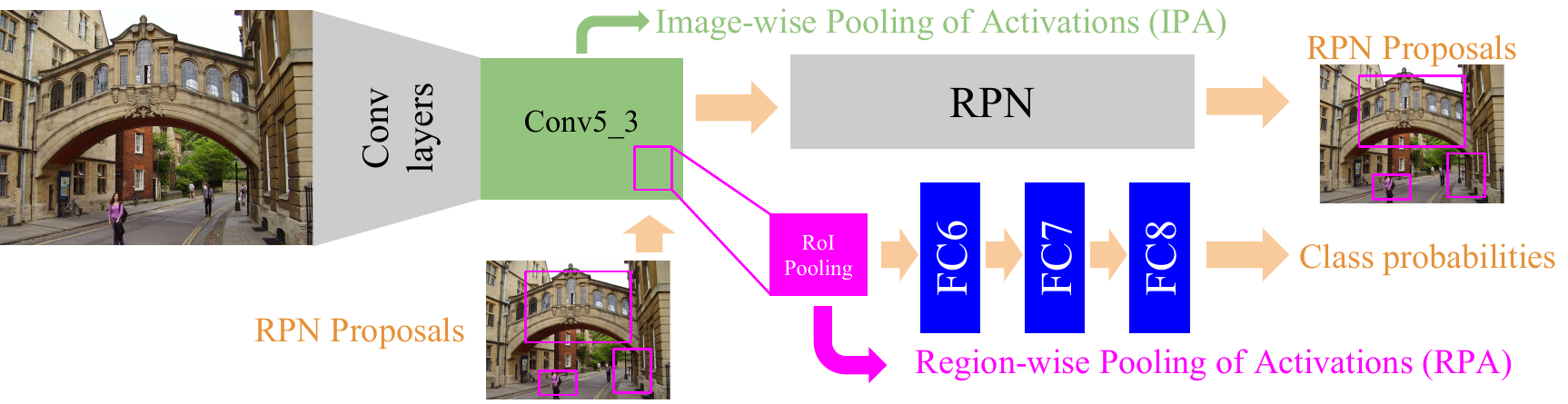}
  \caption{Image- and region-wise descriptor pooling from the Faster R-CNN architecture.}
  \label{fig_frcnn}
\end{figure*}

This way, we put together a simple instance retrieval system that uses both local and global features from an object detection network. Figure \ref{res_examples} shows examples of rankings generated with our retrieval pipeline.

The remainder of the paper is structured as follows. Section \ref{relatedwork} introduces the related works, Section \ref{methodology} presents the methodology of this paper, including feature pooling, reranking and fine-tuning strategies. Section \ref{experiments} includes the performed experiments on three different image retrieval benchmarks as well as the comparison to other state of the art CNN-based instance search systems. Finally, Section \ref{conclusions} draws the conclusions of this work.
\section{Related Work}
\label{relatedwork}

\textbf{CNNs for Instance Search.} Features from pre-trained image classification CNNs have been widely used for instance search in the literature. Early works in this direction demonstrated the suitability of features from fully connected layers for image retrieval \cite{neuralcodes}. Razavian \emph{et al.} \cite{cnnofftheshelf} later improved the results by combining fully connected layers extracted from different image sub-patches. 

A second generation of works explored the usage of other layers in the pretrained CNN and found that convolutional layers significantly outperformed fully connected ones at image retrieval tasks \cite{razavian2015}. Babenko and Lempitsky \cite{babenko2015} later proposed a compact descriptor composed of the sum of the activations of each of the filter responses in a convolutional layer. Tolias \emph{et al.} introduced R-MAC \cite{tolias2015}, a compact descriptor composed of the aggregation of multiple region features. Kalantidis \emph{et al.} \cite{kalantidis2015} found significant improvements when applying non-parametric spatial and channel-wise weighting strategies to the convolutional layers. 

This work shares similarities with all the former in the usage of convolutional features of a pretrained CNN. However, we choose to use a state-of-the-art object detection CNN, to extract both image- and region-based convolutional features in a single forward pass.

\textbf{Object Detection CNNs.} Many works in the literature have proposed CNN-based object detection pipelines. Girshick \emph{et al.} presented R-CNN \cite{rcnn}, a version of Krizhevsky's AlexNet \cite{alexnet}, fine-tuned for the Pascal VOC Detection data \cite{pascal}. Instead of full images, the regions of an object proposal algorithm \cite{selectivesearch} were used as inputs to the network. At test time, fully connected layers for all windows were extracted and used to train a bounding box regressor and classifier.

Since then, great improvements to R-CNN have been released, both in terms of accuracy and speed. He \emph{et al.} proposed SPP-net \cite{sppnet}, which used a Spatial Pyramid based pooling layer to improve classification and detection performance. Additionally, they significantly decreased computational time by pooling region features from convolutional features instead of forward passing each region crop through all layers in the CNN. This way, the computation of convolutional features is shared for all regions in an image. Girshick later released Fast R-CNN \cite{fastrcnn}, which used the same speed strategy as SPP-net but, more importantly, replaced the post-hoc training of SVM classifiers and box regressors with an end-to-end training solution. Ren \emph{et al.} introduced Faster R-CNN \cite{fasterrcnn}, which removed the object proposal dependency of former object detection CNN systems by introducing a Region Proposal Network (RPN). In Faster R-CNN, the RPN shares features with the object detection network in \cite{fastrcnn} to simultaneously learn prominent object proposals and their associated class probabilities. 

In this work, we take advantage of the end-to-end self-contained object detection architecture of Faster R-CNN to extract both image and region features for instance search.
\section{Methodology}
\label{methodology}

\subsection{CNN-based Representations}
\label{met_cnnrepresentations}
This paper explores the suitability of using features from an object detection CNN for the task of instance search. In our setup, query instances are defined by a bounding box over the query images. We choose the architecture and pre-trained models of Faster R-CNN \cite{fasterrcnn} and use it as a feature extractor at both global and local scales. Faster R-CNN is composed of two branches that share convolutional layers. The first branch is a Region Proposal Network that learns a set of window locations, and the second one is a classifier that learns to label each window as one of the classes in the training set.

Similarly to other works \cite{babenko2015,tolias2015,kalantidis2015} our goal is to extract a compact image representation built from the activations of a convolutional layer in a CNN. Since Faster R-CNN operates at global and local scales, we propose the following strategies of feature pooling:

\textbf{Image-wise pooling of activations (IPA).} In order to construct a global image descriptor from Faster R-CNN layer activations, one can choose to ignore all layers in the network that operate with object proposals and extract features from the last convolutional layer. Given the activations of a convolutional layer extracted for an image, we aggregate the activations of each filter response to construct an image descriptor of the same dimension as the number of filters in the convolutional layer. Both max and sum pooling strategies are considered and compared in Section \ref{exp_offtheshelf} of this paper.

\textbf{Region-wise pooling of activations (RPA).} After the last convolutional layer, Faster R-CNN implements a region pooling layer that extracts the convolutional activations for each of the object proposals learned by the RPN. This way, for each one of the window proposals, it is possible to compose a descriptor by aggregating the activations of that window in the RoI pooling layer, giving raise to the region-wise descriptors. For the region descriptor, both max and sum pooling strategies are tested as well. 

Figure \ref{fig_frcnn} shows a schematic of the Faster R-CNN architecture and the two types of descriptor pooling described above.

Following several other authors~\cite{babenko2015,kalantidis2015}, sum-pooled features are $l_2$-normalized, followed by whitening and a second round of $l_2$-normalization, while max-pooled features are just $l_2$-normalized once (no whitening).

\subsection{Fine-tuning Faster R-CNN}
\label{met_finetuning}
 This paper explores the suitability of fine-tuning Faster R-CNN to 1) obtain better feature representations for image retrieval and 2) improve the performance of spatial analysis and reranking. To achieve this, we choose to fine tune Faster R-CNN to detect the query objects to be retrieved by our system. This way, we modify the architecture of Faster R-CNN to output the regressed bounding box coordinates and the class scores for each one of the query instances of the tested datasets. 

In our experiments, we explore two modalities of fine-tuning:

\begin{itemize}

\item \textbf{Fine-tuning Strategy \#1}:  Only the weights of the fully connected layers in the classification branch are updated (i.e. the convolutional layers and the RPN are left unchanged).   

\item \textbf{Fine-tuning Strategy \#2}: Weights of all layers after the first two convolutional layers are updated. This way, convolutional features, RPN proposals and fully connected layers are modified and adapted to the query instances.

\end{itemize}

The resulting fine-tuned networks are to be used to extract better image and region representations and to perform spatial reranking based on class scores instead of feature similarities.

\subsection{Image Retrieval}
\label{met_retrieval}

The three stages of the proposed instance retrieval pipeline are described in this section: filtering stage, spatial reranking and query expansion.

\textbf{Filtering Stage.} The Image-wise pooling (IPA) strategy is used to build image descriptors for both query and database images. At test time, the descriptor of the query image is compared to all the elements in the database, which are then ranked based on the cosine similarity. At this stage, the whole image is considered as the query.

\textbf{Spatial Reranking.} After the Filtering Stage, the top N elements are locally analyzed and reranked. We explore two reranking strategies:

\begin{itemize}
\item \emph{Class-Agnostic Spatial Reranking (CA-SR)}. For every image in the top \emph{N} ranking, the region-wise descriptors (RPA) for all RPN proposals are compared to the region descriptor of the query bounding box. The region-wise descriptors of RPN proposals are pooled from the RoI pooling layer of Faster R-CNN (see Figure \ref{fig_frcnn}). To obtain the region-wise descriptor of the query object, we warp its bounding box to the size of the feature maps in the last convolutional layer and pool the activations within its area. The region with maximum cosine similarity for every image in the top \emph{N} ranking gives the object localization, and its score is kept for ranking. 

\item \emph{Class-Specific Spatial Reranking (CS-SR)}. Using a network that has been fine-tuned with the same instances one wishes to retrieve, it is possible to use the direct classification scores for each RPN proposal as the similarity score to the query object. Similarly to CA-SR, the region with maximum score is kept for visualization, and the score is used to rank the image list.
\end{itemize}

\textbf{Query Expasion (QE).} The image descriptors of the top \emph{M} elements of the ranking are averaged together with the query descriptor to perform a new search.

\section{Experiments}
\label{experiments}

\subsection{Datasets}

The methodologies described in Section \ref{methodology} are assessed with the following datasets:

\begin{itemize}
\item \textbf{Oxford Buildings \cite{philbin2007object}}. 5,063 images, including 55 query images of 11 different buildings in Oxford (5 images/instance are provided). A bounding box surrounding the target object is provided for query images.

\item \textbf{Paris Buildings \cite{paris6k}}. 6,412 still images of Paris landmarks, including 55 query images of 11 buildings with associated bounding box annotations.

\item \textbf{INS 2013  \cite{trecvid}}. A subset of 23,614 keyframes from TRECVid Instance Search (INS) dataset containing only those keyframes that are relevant for at least one of the queries of INS 2013. 
\end{itemize}

\subsection{Experimental Setup}

We use both the VGG16 \cite{vgg} and ZF \cite{zf} architectures of Faster R-CNN to extract image and region features. In both cases, we use the last convolutional layer ($conv5$ and $conv5\_3$ for ZF and VGG16, respectively) to build the image descriptors introduced in Section \ref{methodology}, which are of dimension 256 and 512 for the ZF and VGG16 architectures, respectively. Region-wise features are pooled from the RoI pooling layer of Faster R-CNN. Images are re-scaled such that their shortest side is 600 pixels. All experiments were run in an Nvidia Titan X GPU.

\subsection{Off-the-shelf Faster R-CNN features}
\label{exp_offtheshelf}

In this section, we assess the performance of using off-the-shelf features from the Faster R-CNN network for instance retrieval.

\begin{figure*}[ht]
  \includegraphics[width=\textwidth]{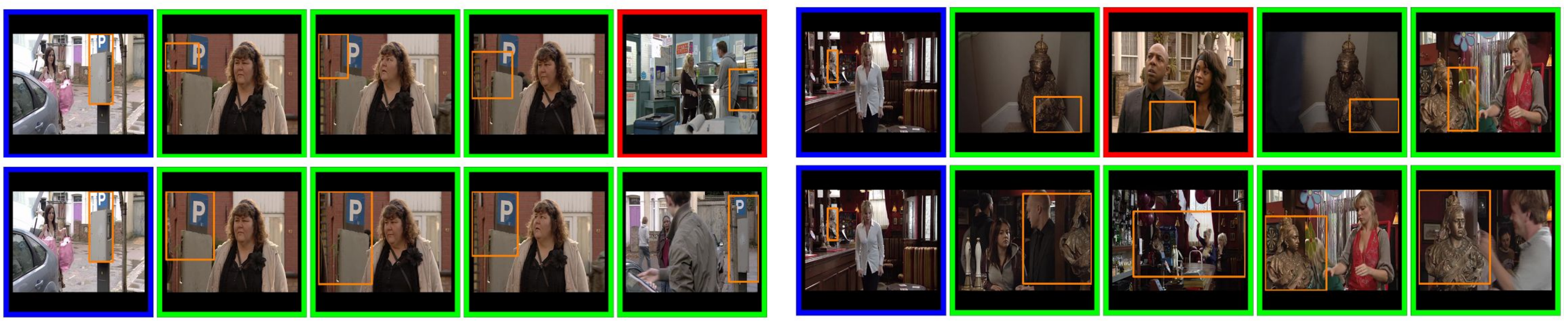}
  \caption{Examples of top 4 rankings and object locations obtained for queries 9098: \emph{a P (parking automat) sign} and 9076: \emph{this monochrome bust of Queen Victoria} from the INS 2013 dataset (query images surrounded in blue). Comparison between the rankings generated using RPA-sum (top) and RPA-max (bottom), after the filtering stage with IPA-sum. Regressed bounding box coordinates have been disabled for visualization.}
  \label{fig:max_sum_comparison}
\end{figure*}

First, we compare the sum and max pooling strategies of image- and region-wise descriptors. Table \ref{max_sum_comparison} summarizes the results. According to our experiments sumpooling is significantly superior to maxpooling for the filtering stage. Such behaviour is consistent with other works in the literature \cite{babenko2015,kalantidis2015}. Sumpooling is, however, consistently outperformed by maxpooling when reranking using region-wise features for all three datasets. Specifically for the Oxford and Paris datasets, we find the spatial reranking with maxpooling to be beneficial after filtering (gain of 0.10 and 0.03 mAP points for Oxford and Paris, respectively). However, the spatial reranking (either with max or sum pooling) has little or no effect for the INS13 dataset. To further interpret these results, we qualitatively evaluate the two pooling strategies. Figure \ref{fig:max_sum_comparison} shows examples of top rankings for INS13 queries, spatially reranked with region-wise max and sum pooled descriptors. These examples indicate that, although mAP is similar, the object locations obtained with maxpooling are more accurate. According to this analysis, we set IPA-sum descriptors for the filtering stage and RPA-max descriptors for the spatial reranking in all the upcoming experiments of this paper.

Table \ref{frcnn_ots_qe} shows the performance of different Faster R-CNN architectures (ZF and VGG16) trained on two datasets (Pascal VOC and COCO \cite{coco}), including experiments with query expansion with the $M = 5$ top retrieved images as well. As expected, features pooled from the deeper VGG16 network perform better in most cases, which is consistent with previous works in the literature showing that features from deeper networks reach better performance. Query expansion applied after the spatial reranking achieves significant gains for all tested datasets. Such behaviour was expected in particular with Oxford and Paris datasets, for which the spatial reranking already provided a significant gain. Interestingly, query expansion is also most beneficial after spatial reranking for the INS13 dataset, which suggests that, although in this case the spatial reranking does not provide any gain in mAP, the images that fall on the very top of the ranking are more useful to expand the query than the ones in the top of the first ranking.

\begin{table}[]
\centering
\caption{Mean Average Precision (mAP) comparison between sum and max pooling strategies for both filtering and reranking stages using \emph{conv5} features from the ZF Faster R-CNN model.}
\label{max_sum_comparison}
\begin{tabular}{@{}ccccc@{}}
\toprule
Filtering                & Reranking & Oxford 5k & Paris 6k & INS 13 \\ \midrule
\multirow{3}{*}{IPA-sum} & None      & 0.505     & 0.612    & \textbf{0.215}  \\
                         & RPA-sum   & 0.501     & 0.621    & 0.196  \\
                         & RPA-max   & \textbf{0.602}     & \textbf{0.641}    & 0.206  \\ \midrule
\multirow{3}{*}{IPA-max} & None      & 0.478     & 0.540    & 0.131  \\
                         & RPA-sum   & 0.508     & 0.565    & 0.135  \\
                         & RPA-max   & 0.559     & 0.561    & 0.138  \\ \midrule 
\end{tabular}
\end{table}

\begin{table}[]
\small
\centering
\caption{mAP of pre-trained Faster R-CNN models with ZF and VGG16 architectures. (P) and (C) denote whether the network was trained with Pascal VOC or Microsoft COCO images, respectively. In all cases, IPA-sum descriptors are used for the filtering stage.  The CA-SR column specifies whether Class-Agnostic Spatial Reranking with RPA-max is applied to the top $N = 100$ elements of the ranking. When indicated, QE is applied with $M = 5$.}
\label{frcnn_ots_qe}
\begin{tabular}{@{}cccccc@{}}
\toprule
Net                       & CA-SR                    & QE  & Oxford 5k      & Paris 6k       & INS 13         \\ \midrule
\multirow{4}{*}{ZF (P)}    & \multirow{2}{*}{No}  & No  & 0.505          & 0.612          & 0.215          \\
                          &                      & Yes & 0.515          & 0.671          & 0.246          \\
                          & \multirow{2}{*}{Yes} & No  & 0.602          & 0.640          & 0.206          \\ 
                          &                      & Yes & 0.622          & 0.707          & \textbf{0.261} \\ \midrule
\multirow{4}{*}{VGG16 (P)} & \multirow{2}{*}{No}  & No  & 0.588          & 0.657          & 0.172          \\
                          &                      & Yes & 0.614          & 0.706          & 0.201          \\
                          & \multirow{2}{*}{Yes} & No  & 0.641          & 0.683          & 0.171          \\
                          &                      & Yes & \textbf{0.679} & 0.729          & 0.242          \\ \midrule
\multirow{4}{*}{VGG16 (C)} & \multirow{2}{*}{No}  & No  & 0.588          & 0.656          & 0.216          \\
                          &                      & Yes & 0.600          & 0.695          & 0.250          \\
                          & \multirow{2}{*}{Yes} & No  & 0.573          & 0.663          & 0.192          \\
                          &                      & Yes & 0.647          & \textbf{0.732} & 0.241          \\ \midrule 
\end{tabular}
\end{table}

\subsection{Fine-tuning Faster R-CNN}

In this section, we assess the impact in retrieval performance of fine-tuning a pretrained network with the query objects to be retrieved. We choose to fine-tune the VGG16 Faster R-CNN model, pretrained with the objects of the Microsoft COCO dataset. 

In the case of Oxford and Paris, we modify the output layers in the network to return 12 class probabilities (11 buildings in the dataset, plus an extra class for the background), and their corresponding regressed bounding box coordinates. We use the 5 images provided for each one of the buildings and their bounding box locations as training data. Additionally, we augment the training set by performing a horizontal flip on the training images ($11*5*2 = 110$ training images in total). For INS 13, we have 30 different query instances, with 4 images each, giving raise to $30*4*2 = 240$ training examples. The number of output classes for INS 13 is 31 (30 queries plus the background class).

The original Faster R-CNN training parameters described in \cite{fasterrcnn} are kept for fine-tuning, except for the number of iterations, which we decreased to 5.000 considering our small number of training samples. We use the approximate joint training strategy introduced in \cite{fasterrcnn-arxiv}, which trains the RPN and classifier branches at the same time, using the multi-task loss defined in \cite{fasterrcnn}. This way, we train a separate network for each one of the tested datasets, using the two different fine-tuning modalities described in Section \ref{met_finetuning}. Fine-tuning was performed on a Nvidia Titan X GPU and took around 30 and 45 minutes for finetuning strategies \#1 and \#2, respectively.

We first take the networks fine-tuned with strategy \#1 and run the retrieval pipeline from scratch. Table \ref{ft_exp} shows the obtained results (ft\#1 columns). Results of the filtering and CA-SR stages are the same as those obtained with the original Faster R-CNN model, which is because the weights for the convolutional layers were not modified during fine-tuning. Results indicate that, although mAP is not always improved after CS-SR (e.g. from 0.588 to 0.543 for Oxford 5k), it is significantly better than CA-SR for Oxford and Paris when followed with query expansion. In case of the INS 13 dataset, we do not find significant improvements when using CS-SR, which suggests that only fine-tuning fully connected layers might not be sufficient to effectively detect the challenging query objects in this dataset.

\begin{figure}
  \includegraphics[width=\columnwidth]{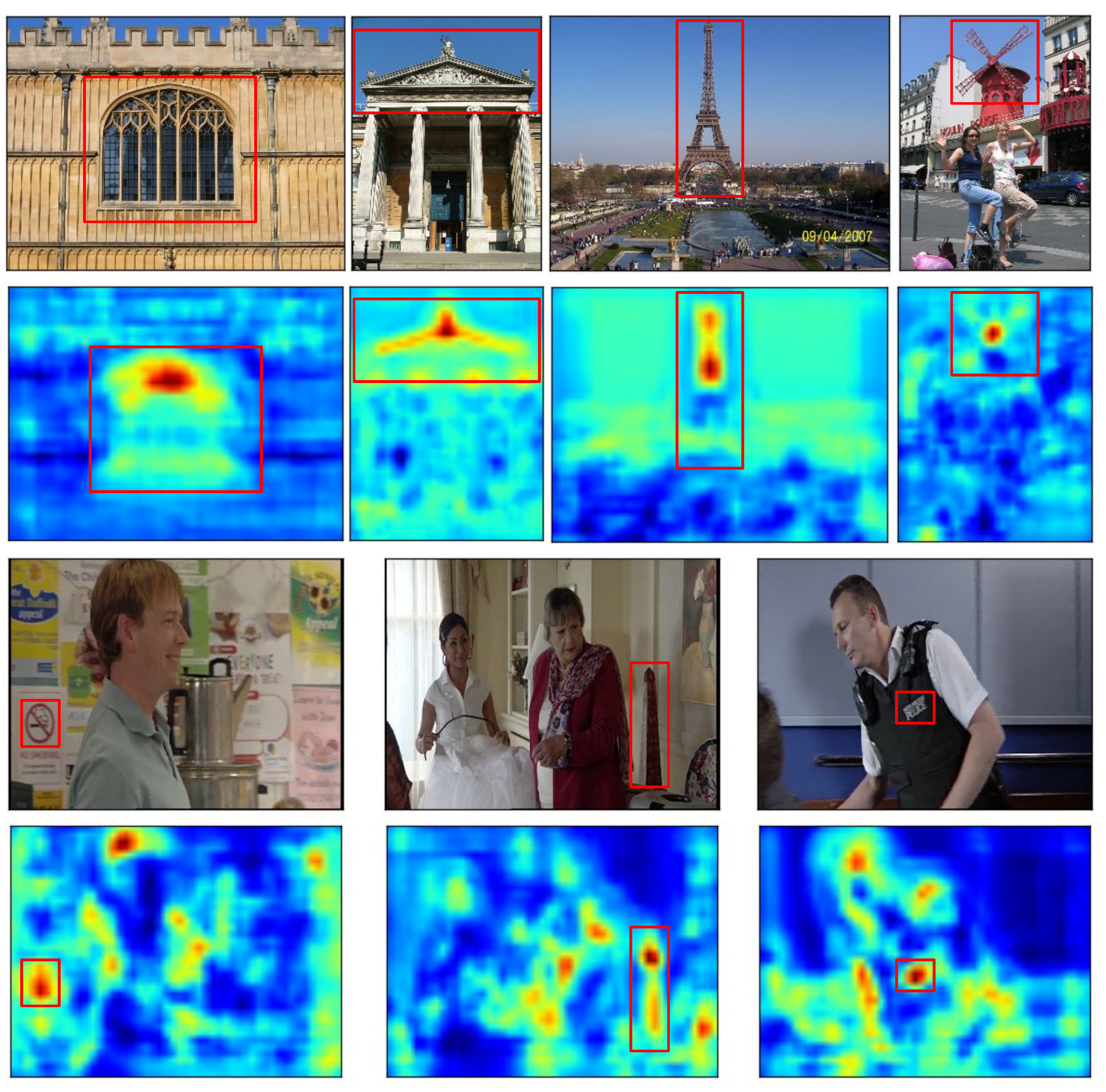}
  \caption{Difference between conv5\_3 features (sum pooled over feature maps) extracted from the original Faster R-CNN model pretrained with MS COCO with conv5\_3 features from the same model fine-tuned for INS13 (bottom), Oxford and Paris (top) queries.}
  \label{fig_diff}
\end{figure}

The second experiment in this section involves fine-tuning a higher number of layers in the Faster R-CNN architecture (Fine-tuning Strategy \#2). Using this modality, the weights in the last convolutional layer are modified. Figure \ref{fig_diff} shows the difference in the activations in conv5\_3 after fine-tuning it for the query instances in each dataset. These visualizations indicate that, after fine-tuning, more neurons in the convolutional layer positively react to the visual patterns that are present in the query objects of the dataset.

We then use the fine-tuned networks of the Fine-tuning Strategy \#2 for each one of the datasets to extract image- and region-wise descriptors to perform instance search. Table \ref{ft_exp} presents the results (ft\#2 columns). As expected, fine-tuned features significantly outperform raw Faster R-CNN features for all datasets (mAP is $\sim$ 20\% higher for Oxford and Paris, and 8\% higher for INS 13). Results indicate that, for Oxford and Paris datasets, the gain of CA-SR + QE is higher with raw features (10\% and 11\% mAP increase for Oxford and Paris, respectively) than with fine-tuned ones (8\% and 3\% mAP increase, respectively). This suggests that fine-tuned features are already discriminant enough to correctly retrieve the objects in these two datasets. However, results for the INS 13 dataset show that CA-SR + QE is most beneficial when using fine-tuned features (11\% and 41\% mAP increase for raw and fine-tuned features, respectively). This difference between the performance for Oxford/Paris and INS13 suggests that queries from the latter are more challenging and therefore benefit from fine-tuned features and spatial reranking the most. A similar behaviour is observed for CS-SR which, for Oxfod and Paris, is most beneficial when applied to a ranking obtained with raw features. For INS 13, however, the gain is greater when using fine-tuned features. Overall, the performance of reranking + query expansion is higher for CS-SR than CA-SR. Figure \ref{res_examples} shows examples of rankings for queries of the three different datasets after applying CS-SR. For visualization, we disable the regressed bounding box coordinates predicted by Faster R-CNN and choose to display those that are directly returned by the RPN. We find that the locations returned by the regression layer are innacurate in most cases, which we hypothesize is caused by the lack of training data.

\begin{figure*}
  \includegraphics[width=\textwidth]{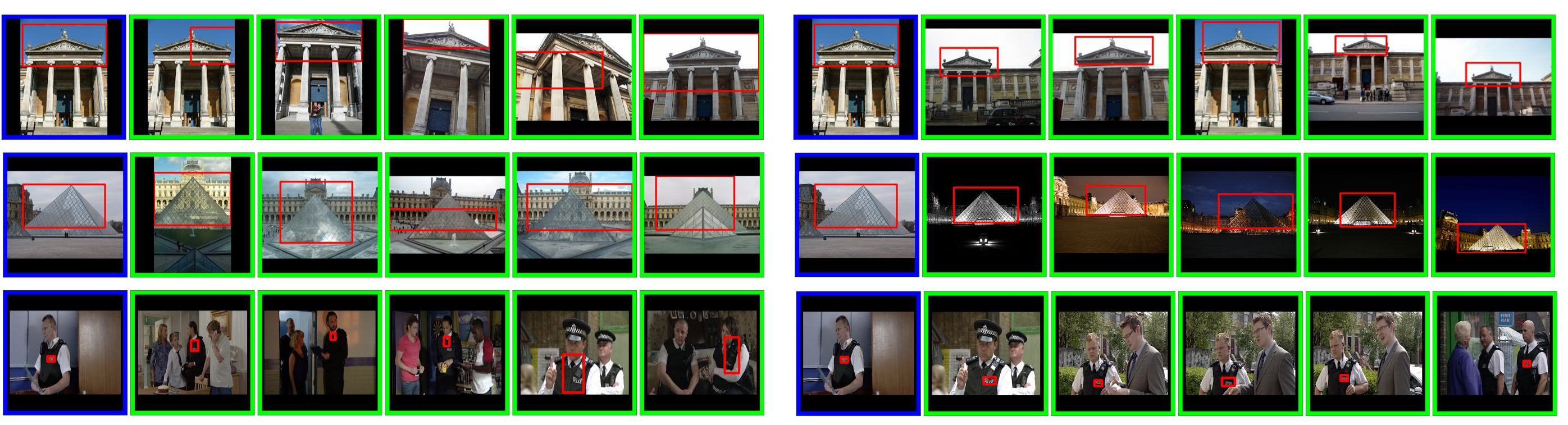}
  \caption{Ranking examples after CS-SR with fine-tuned strategies \#1 (left) and \#2 (right).}
  \label{ft_diff}
\end{figure*}

Finally, in Figure \ref{ft_diff} we qualitatively evaluate the object detections after CS-SR using the fine-tuned strategies \#1 and \#2. The comparison reveals that locations obtained with the latter are more accurate and tight to the objects. The Fine-tuning Strategy \#2 allows the RPN layers to adapt to the query objects, which causes the network to produce object proposals that are more suitable for the objects in the test datasets.

\begin{table}[]
\centering
\small
\caption{Comparison between Fine-tuning strategies \#1 (ft\#1) and \#2 (ft\#2) on the three datasets. Spatial Reranking (R) is applied to the $N = 100$ top elements of the ranking. QE is performed with $M = 5$.}
\label{ft_exp}
\begin{tabular}{@{}cccccccc@{}}
\toprule
R & QE  & \multicolumn{2}{c}{Oxford 5k} & \multicolumn{2}{c}{Paris 6k} & \multicolumn{2}{c}{INS 13} \\ \midrule
          &     & ft\#1           & ft\#2            & ft\#1            & ft\#2           & ft\#1              & ft\#2       \\
No        & No  & 0.588           & 0.710            & 0.656            & 0.798           & 0.216              & 0.234       \\
No        & Yes & 0.600           & 0.748            & 0.695            & 0.813           & \textbf{0.250}     & 0.259       \\
CA-SR       & No  & 0.573           & 0.739            & 0.663            & 0.801           & 0.192              & 0.248       \\
CA-SR       & Yes & 0.647           & 0.772            & 0.732            & 0.824           & 0.241              & 0.330       \\
CS-SR     & No  & 0.543           & 0.751            & \textbf{0.793}   & 0.807           & 0.181              & 0.250       \\
CS-SR     & Yes &\textbf{0.678}   & \textbf{0.786}   & 0.784            & \textbf{0.842}  & \textbf{0.250}     & \textbf{0.339}       \\ \bottomrule
\end{tabular}
\end{table}

\subsection{Comparison with state-of-the-art}

In this section, we compare our results with several instance search works in the literature. Table \ref{soa} shows the results of this comparison. 

Our proposed pipeline using Faster R-CNN features shows competitive results with respect to the state of the art. However, other works \cite{kalantidis2015,tolias2015} achieve a very high performance without any reranking nor query expansion strategies using similar feature pooling strategies. We hypothesize that the difference in the CNN architecture (Faster R-CNN vs. VGG16), training data (Pascal VOC vs ImageNet) and input image size (600px wide vs. full resolution) between these works and ours might be the reasons of the gap in performance. Our proposed reranking strategy CA-SR followed by query expansion is demonstrated to provide similar mAP gains compared to the one proposed in \cite{tolias2015}. While CA-SR + QE gives us a gain in mAP of $\sim$ 10\% both for Oxford and Paris (using raw Faster R-CNN features), Tolias \emph{et al.} \cite{tolias2015} use their reranking strategy to raise their mAP by 5 and 15\% for the two datasets, respectively.     

As expected, results obtained with fine-tuned features (ft\#2) achieve very competitive results compared to those in the state of the art, which suggests that fine-tuning the network for the object queries is an effective solution when time is not a constraint. 

\begin{table}[]
\centering
\caption{Comparison with CNN-based state-of-the-art works on instance retrieval.}
\label{soa}
\begin{tabular}{@{}lcc@{}}
\toprule
                            & Oxford 5k      & Paris 6k       \\ \midrule
Razavian \emph{et al.} \cite{cnnofftheshelf}            & 0.556          & 0.697          \\
Tolias \emph{et al.} \cite{tolias2015}               & 0.668          & \textbf{0.830}          \\
Kalantidis \emph{et al.} \cite{kalantidis2015}           & 0.682          & 0.796          \\
Babenko and Lempitsky \cite{babenko2015}             & 0.657          & -              \\
Ours                        & 0.588          & 0.656          \\
Ours (ft\#2)                & \textbf{0.710}          & 0.798          \\ \midrule
Tolias \emph{et al.} (+ R + QE) \cite{tolias2015}   & 0.770          & \textbf{0.877}          \\
Kalantidis \emph{et al.} (+ QE) \cite{kalantidis2015}  & 0.722          & 0.855          \\
Ours (+ CA-SR + QE)           & 0.647          & 0.732          \\
Ours (ft\#1) (+ CS-SR + QE) & 0.678 & 0.784 \\
Ours (ft\#2) (+ CS-SR + QE) & \textbf{0.786} & 0.842 \\\bottomrule
\end{tabular}
\end{table}

\section{Conclusion}
\label{conclusions}

This paper has presented different strategies to make use of CNN features from an object detection CNN. It provides a simple baseline that uses off-the-shelf Faster R-CNN features to describe both images and their sub-parts. We have shown that is possible to greatly improve the performance of an off-the-shelf based system, at the cost of fine tuning the CNN for the query images that include objects that one wants to retrieve.
\section*{Acknowledgements}

This work has been developed in the framework of the project BigGraph TEC2013-43935-R, funded by the Spanish Ministerio de Econom\'ia y Competitividad and the European Regional Development Fund (ERDF). The Image Processing Group at the UPC is a SGR14 Consolidated Research Group recognized and sponsored by the Catalan Government (Generalitat de Catalunya) through its  AGAUR office. Amaia Salvador developed this work thanks to the NII International Internship Program 2015.
We gratefully acknowledge the support of NVIDIA Corporation with the donation of the GeForce GTX Titan X used in this work.

{\small
\bibliographystyle{ieee}
\bibliography{egbib}
}

\end{document}